\documentclass[times,preprint,10pt,numbers]{elsarticle}

\usepackage[numbers]{natbib}
\usepackage{amsmath}
\usepackage{amssymb}
\usepackage{bm}
\usepackage{booktabs}
\usepackage{multirow}
\usepackage{makecell}
\usepackage{tabularx}
\usepackage{graphicx}
\usepackage{colortbl}
\usepackage{nicematrix}
\usepackage{wrapfig}
\usepackage[colorlinks=true,linkcolor=blue,citecolor=blue,urlcolor=blue]{hyperref}
\usepackage{xcolor}
\definecolor{mosaicrow}{RGB}{225,244,252}

\newcolumntype{L}{@{\extracolsep{\fill}}l}
\newcolumntype{R}{@{\extracolsep{\fill}}r}
\newcolumntype{C}{@{\extracolsep{\fill}}c}

\begin{document}

\begin{frontmatter}

\title{Motion-Saliency Complementary Masked Modeling for Point Cloud Video Understanding}

\author[1,2]{Wei Wang\fnmark[fn1]}\ead{wangwei2121@qq.com}
\author[3,5]{Yiding Sun\fnmark[fn1]}\ead{sunyiding@stu.xjtu.edu.cn}
\author[3]{Yuyan Wang\fnmark[fn1]}\ead{wangyy0718@stu.xjtu.edu.cn}
\author[3]{Zhuoyue Zhang}\ead{zhangzy4016@stu.xjtu.edu.cn}
\author[4]{Zhengqiao Li}\ead{ZQlzhengqiao@stu.xjtu.edu.cn}

\cortext[cor1]{Corresponding authors.}

\author[5]{\corref{cor1}Dongfu Yin}\ead{yindongfu@gml.ac.cn}
\author[1]{\corref{cor1}Chen Li}\ead{cli@xjtu.edu.cn}

\fntext[fn1]{These authors contributed equally to this work.}

\affiliation[1]{organization={School of Computer Science and Technology, Xi'an Jiaotong University},
            city={Xi'an},
            postcode={710049},
            country={China}}

\affiliation[2]{organization={State Key Laboratory of Intelligent Geotechnics and Tunnelling (FSDI)},
            city={Xi'an},
            postcode={710043},
            country={China}}

\affiliation[3]{organization={School of Software Engineering, Xi'an Jiaotong University},
            city={Xi'an},
            postcode={710049},
            country={China}}

\affiliation[4]{organization={School of Information and Communication Engineering, Xi'an Jiaotong University},
            city={Xi'an},
            postcode={710049},
            country={China}}

\affiliation[5]{organization={Guangdong Laboratory of Artificial Intelligence and Digital Economy (SZ)},
            city={Shenzhen},
            postcode={518107},
            country={China}}

\begin{abstract}
Point cloud video representation learning is crucial for 3D dynamic scene understanding. In this paper, we propose MoSaiC, a novel Motion-Saliency Complementary masked modeling framework for self-supervised point cloud video representation learning. MoSaiC couples three components, Curriculum Motion-Saliency Masking (CMSM), which guides the masking process toward motion-salient tokens under a curriculum schedule, Normal-Flow Motion (NFM) modeling, which supervises the local rigid rotation of each token in the Lie algebra $\mathfrak{so}(3)$ as an explicit geometric motion target, and Cross-view Token Consistency Prediction (CTCP), which enforces consistency between two complementary masked views at the token level. Together, these components allow MoSaiC to effectively capture both appearance and motion dynamics. Extensive experiments on multiple downstream tasks, including action recognition, temporal action segmentation, and point-level semantic segmentation, demonstrate the effectiveness of our approach.
\end{abstract}


\end{frontmatter}

\section{Introduction}\label{sec,intro}
Dynamic 3D perception seeks to understand not only the geometry of a scene, but also how that geometry evolves over time. Point cloud videos, sequences of 3D point sets captured by depth cameras or LiDAR, provide a direct and illumination-robust representation for this purpose. They have enabled progress in action recognition, temporal action segmentation, and 4D semantic segmentation \citep{fan2021p4transformer, fan2022pst, liu2022hoi4d}, with applications ranging from gesture interfaces to embodied perception and human--object interaction. Their success, however, has largely relied on supervised learning. Annotating a dynamic point cloud requires labels to be assigned across frames, and dense tasks further demand temporal boundaries or point-level semantics. Such annotation becomes prohibitively expensive as sequence length and scene complexity increase.

Self-supervised learning (SSL) offers a way to reduce this dependence by learning transferable spatio-temporal representations from unlabeled point cloud videos. The central problem is not simply how to reconstruct missing 3D content, but how to design a pretext task that forces the encoder to discover the sparse, localized geometric changes that define motion. Unlike RGB videos, point cloud sequences have no persistent pixel grid, their sampling density varies across frames, and point-wise correspondences are generally unavailable. A useful objective must therefore separate informative motion from static geometry without assuming reliable tracking, while retaining the token-level structure required by dense downstream tasks.

Existing point cloud video SSL methods can be grouped into three broad families. \emph{Masked reconstruction methods} hide point tubes and recover their geometry or temporal statistics, as exemplified by MaST-Pre \citep{shen2023mastpre}. \emph{Contrastive and distillation methods} learn invariance between augmented, complete, or partial observations, including PointCMP \citep{shen2023pointcmp}, C2P \citep{zhang2023c2p}, and Query-CP \citep{querycp2024}. More recent \emph{hybrid methods} combine reconstruction with motion or semantic objectives, M2PSC \citep{han2024m2psc}, for example, predicts point trajectories while contrasting masked and visible tubes. These approaches establish the value of unlabeled 4D data, but they still treat the most consequential design choices only partially.

\begin{wrapfigure}{r}{0.48\linewidth}
  \centering
  \includegraphics[width=\linewidth]{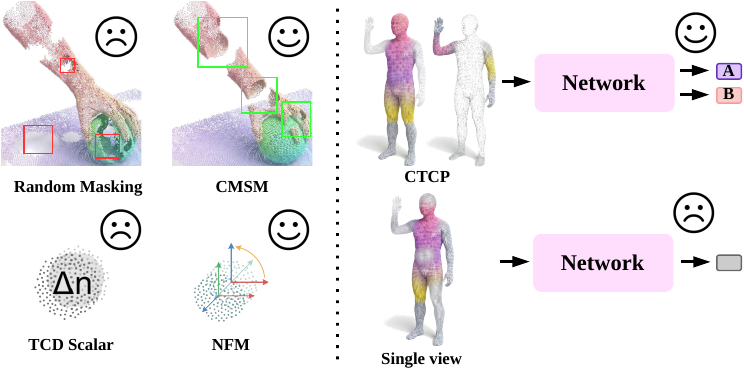}
  \caption{Motivation of MoSaiC. Existing point cloud video SSL methods suffer from motion-agnostic masking, coarse or implicit motion modeling, and single-view reconstruction. MoSaiC addresses them with Curriculum Motion-Saliency Masking (CMSM), Normal-Flow Motion (NFM) modeling, and Cross-view Token Consistency Prediction (CTCP).}
  \label{fig,motivation}
\end{wrapfigure}

We identify three coupled limitations, illustrated in Figure~\ref{fig,motivation}. First, existing masking distributions are random or otherwise independent of local geometric change, so scarce reconstruction capacity is spent equally on static and dynamic tokens. Second, motion is represented by coarse statistics, implicit feature objectives, or correspondence-dependent trajectories. A scalar cardinality change cannot distinguish density-preserving rotations, while implicit motion features do not specify the axis and angle of local reorientation. Third, single-view reconstruction and tube- or clip-level contrast provide no direct consistency constraint for the same token under complementary partial observations. Consequently, existing objectives do not jointly determine \emph{where} motion should be learned, \emph{what} geometric motion should be encoded, and \emph{how} local representations should remain stable across views.

These limitations directly affect downstream performance. Motion-agnostic masking makes a model proficient at interpolating redundant background geometry while undersupervising the few tokens that distinguish similar actions. Coarse motion targets merge physically different transformations, weakening the cues needed to separate rotation-heavy actions and to localize transitions between motion states. Global or single-view objectives can still produce useful clip descriptors, but they leave individual token features insufficiently constrained for temporal and point-level prediction. Addressing any one limitation in isolation is also inadequate, focusing on dynamic tokens before learning basic geometry makes the reconstruction task prematurely difficult, whereas a precise motion target has limited value if it is rarely applied to informative regions.

To address this problem, we propose \textbf{MoSaiC}, a Motion-Saliency Complementary masked modeling framework that couples three components. \emph{Curriculum Motion-Saliency Masking (CMSM)} estimates token saliency from temporal SHOT differences and gradually changes the masking distribution from uniform to motion-focused, so the encoder learns a geometric prior before being challenged with highly dynamic regions. \emph{Normal-Flow Motion (NFM)} supplies an explicit, correspondence-free motion target by regressing the relative rotation between adjacent local PCA reference frames in the Lie algebra $\mathfrak{so}(3)$, while a Temporal Descriptor Difference (TDD) target preserves complementary cues from deformation and occlusion. \emph{Cross-view Token Consistency Prediction (CTCP)} constructs complementary masked views and aligns the decoded feature of a hidden token with its encoded feature when that token is visible in the other view. In this way, CMSM selects \emph{where} to learn, NFM and TDD define \emph{what} motion to capture, and CTCP regularizes \emph{how} the resulting token representation behaves across partial observations.

Our experiments also reveal several results that are not obvious from standard masked-modeling practice. Masking motion-salient tokens from the beginning is inferior to introducing them through a curriculum, showing that harder masking is beneficial only after a basic geometric representation has formed. A mask ratio of $0.6$ outperforms the commonly used higher ratios because excessively masking one view makes its complementary view nearly trivial. More importantly, although motion supervision might appear irrelevant or even detrimental to static point-level labeling, MoSaiC improves HOI4D semantic segmentation. Its largest action-segmentation gains occur for rotation-dominated classes and at the strictest temporal overlap threshold, suggesting that explicit local orientation modeling preserves fine surface geometry while sharpening motion boundaries.

The main contributions of this paper are summarized as follows,
\begin{itemize}
\item We formulate motion-aware point cloud video SSL around three coupled requirements, motion-targeted masking, explicit geometric motion supervision, and token-level cross-view consistency, exposing why treating these requirements separately leaves important spatio-temporal cues underused.
\item We develop MoSaiC, which integrates CMSM, correspondence-free NFM and complementary TDD targets, and CTCP in a unified masked-modeling framework. The three components provide distinct supervision over target selection, local motion geometry, and cross-view token prediction.
\item Extensive experiments on MSR-Action3D, SHREC'17, NvGesture, and HOI4D demonstrate transferable and label-efficient gains in action recognition, temporal action segmentation, and semantic segmentation. Ablations explain the counter-intuitive effects of curriculum masking, moderate mask ratios, and motion supervision on static semantic prediction, while downstream inference retains only the backbone and incurs no additional cost.
\end{itemize}

\section{Related Work}\label{sec,related}
\subsection{Point Cloud Video Modeling}
Supervised point cloud video backbones extend static point networks \citep{qi2017pointnet, qi2017pointnetpp} to the temporal domain. MeteorNet \citep{liu2019meteornet} aggregates spatio-temporal neighbors, whereas PSTNet \citep{fan2021pstnet} decouples spatial and temporal convolutions hierarchically. P4Transformer \citep{fan2021p4transformer} embeds local 4D structures through point 4D convolution followed by self-attention, avoiding explicit point tracking, and PST-Transformer \citep{fan2022pst} further searches for related points across a sequence with decoupled spatio-temporal encoding. Kinet \citep{zhong2022kinet} models feature-level dynamics by fitting space-time surfaces, and PPTr \citep{wen2022pptr} uses primitive planes to capture long-range context \citep{zhang2026cmhanet, zhang2026pointcot, sun2026pointata, guo2026mantis, wang2026pointrft, guo2026parameter, sun2026spikingmot}. These architectures establish effective downstream encoders, but they are trained with costly dense supervision and do not themselves prescribe how motion-aware, label-free objectives should be constructed. MoSaiC retains a P4Transformer-style tube embedding while learning its representation from unlabeled videos through motion-targeted pre-training.

\subsection{Masked and Contrastive Self-Supervised Learning}
Masked autoencoders (MAE) \citep{he2022masked} reconstruct randomly masked image patches with an asymmetric encoder-decoder, and VideoMAE \citep{tong2022videomae} transfers this principle to videos through high-ratio tube masking. For static point clouds, Point-BERT \citep{yu2022pointbert}, Point-MAE \citep{pang2022pointmae}, and Point-M2AE \citep{zhang2022pointm2ae} adapt masked modeling through discrete tokenization, coordinate reconstruction, and hierarchical multi-scale encoding, respectively \citep{sun2026hyperpoint, li2025pointdico, sun2026curve3d, han2025ggm, you2026gaussfusion}. These approaches show that reconstruction can provide transferable geometric priors, but random masking and appearance-centric targets do not distinguish a temporally informative change from a static, trivially reconstructible region. This motivates CMSM and NFM, which direct masking toward motion-salient tokens and explicitly supervise their local geometric motion \citep{zhang2026chain}.

Beyond reconstruction, contrastive methods such as PointContrast \citep{xie2020pointcontrast} contrast features of corresponding points across views, while non-contrastive siamese frameworks such as BYOL \citep{grill2020byol} and SimSiam \citep{chen2021simsiam} achieve view consistency without negative samples. Their targets are generally global descriptors, correspondence-dependent point features, or 2D video features, and therefore do not provide a token-level cross-view constraint for partially observed 4D point sequences. CTCP builds on the asymmetric predictor and stop-gradient design of siamese learning, but applies it only when a token is masked in one complementary view and visible in the other.

\subsection{Self-Supervised Learning on Point Cloud Videos}
A growing body of work studies SSL specifically for point cloud videos. MaST-Pre \citep{shen2023mastpre} performs spatio-temporal point-tube masking and jointly reconstructs masked tubes and predicts temporal cardinality difference (TCD). PointCMP \citep{shen2023pointcmp} unifies contrastive learning and mask prediction, C2P \citep{zhang2023c2p} distills representations from complete sequences to partial ones, and Query-CP \citep{querycp2024} combines learnable query contrast with decoupled spatio-temporal prediction \citep{zhang2026dimp, sun2026pointtrie}. M2PSC \citep{han2024m2psc} is the most closely related prior method, coupling motion trajectory prediction with semantic contrast between visible and masked tubes alongside appearance reconstruction. Its motion target is a per-point trajectory under an implicit correspondence rather than the correspondence-free $\mathfrak{so}(3)$ rotation of NFM, and its contrast operates between masked and visible tubes rather than through CTCP's complementary two-view predictor.

These methods leave the three requirements of Section~\ref{sec,intro} only partially addressed. Their masking distributions are independent of geometric motion, none supplies an explicit rotation target, and none imposes token-level consistency across complementary masked observations. MoSaiC addresses them jointly: CMSM selects motion-salient masking targets through a curriculum, NFM supplies correspondence-free $\mathfrak{so}(3)$ supervision while TDD retains a complementary topology-sensitive signal, and CTCP enforces cross-view consistency at token granularity.

\section{Method}\label{sec,method}
\subsection{Overview}
Figure~\ref{fig,framework} illustrates the overall framework of MoSaiC. Given an input point cloud video, we first tokenize it into a sequence of spatio-temporal tokens. Instead of using a random masking strategy, we propose Curriculum Motion-Saliency Masking (CMSM) to generate two complementary masked views (View A and View B) based on the motion saliency of each token. The visible tokens from both views are fed into a shared asymmetric encoder-decoder architecture. The decoder reconstructs the masked tokens, guided by three types of reconstruction targets, appearance (Chamfer Distance), Normal-Flow Motion (NFM), and Temporal Descriptor Difference (TDD). Furthermore, we introduce Cross-view Token Consistency Prediction (CTCP) to enforce the consistency between the predicted representations of the two complementary views.

\begin{figure*}[t]
  \centering
  \includegraphics[width=\textwidth]{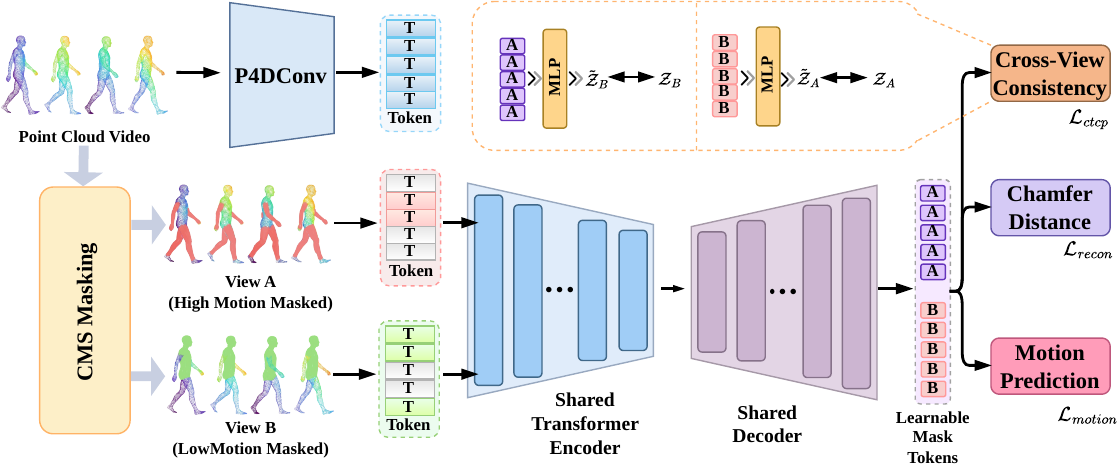}
  \caption{Overview of the MoSaiC pre-training pipeline. CMSM produces two complementary masked views from motion-saliency scores. A shared encoder--decoder reconstructs masked tokens under appearance, NFM, and TDD losses. CTCP aligns the decoded feature of a token hidden in one view with its encoded feature when visible in the other. Only the encoder is retained for downstream tasks.}
  \label{fig,framework}
\end{figure*}

\subsection{Preliminaries, Point Cloud Video Tokenization}\label{sec,prelim}
Given a point cloud video sequence $\mathcal{P} = \{P_1, P_2, \dots, P_T\}$, where $P_t \in \mathbb{R}^{N \times 3}$ is the point cloud frame at time step $t$, we extract spatio-temporal tokens using a tube embedding module based on point 4D convolution (P4DConv) \citep{fan2021p4transformer}, anchor points are sampled via Farthest Point Sampling and grouped with their spatial and temporal neighbors via ball query (hyperparameter values are given in Section~\ref{sec,experiments}). Each resulting token $i$ carries a feature vector $\bm{f}_i \in \mathbb{R}^C$, a spatio-temporal position $\bm{p}_i = (x_i, y_i, z_i, t_i) \in \mathbb{R}^4$, and its local neighborhood points $\mathcal{N}_i$, which span a set of consecutive frames $\mathcal{T}_i$.

For every frame $t \in \mathcal{T}_i$ we additionally compute a Signature of Histograms of OrienTations (SHOT) descriptor $\bm{h}_{i,t} \in \mathbb{R}^L$ over the neighborhood points of token $i$ in that frame \citep{tombari2010shot}. SHOT is conventionally made rotation-invariant by expressing each neighborhood in its own local reference frame, we instead express $\bm{h}_{i,t}$ and $\bm{h}_{i,t+1}$ in a \emph{common} reference frame, the frame $\bm{R}_t$ estimated at time $t$ (Section~\ref{sec,targets}), so that their difference reflects a change of orientation instead of being cancelled by SHOT's built-in invariance. The neighborhood points and these descriptors serve as reconstruction targets for the self-supervised objectives introduced below.

\subsection{Curriculum Motion-Saliency Masking (CMSM)}\label{sec,cmsm}
To encourage the model to focus on dynamic regions, we propose Curriculum Motion-Saliency Masking (CMSM), whose pipeline, from per-token SHOT descriptors to the two complementary masked views, is illustrated in Figure~\ref{fig,cmsm}. We compute the motion saliency $s_i$ of each token $i$ as the mean absolute temporal difference of its SHOT descriptors over the frames its tube spans,

\begin{equation}
    s_i = \frac{1}{L (|\mathcal{T}_i| - 1)} \sum_{t \in \mathcal{T}_i \setminus \{t_{\max}\}} \sum_{l=1}^L \left| \bm{h}_{i, t+1}^{(l)} - \bm{h}_{i, t}^{(l)} \right|,
\end{equation}
where $L$ is the dimension of the SHOT descriptor and $t_{\max}$ is the last frame of $\mathcal{T}_i$. Because the raw scores occupy a narrow, clip-dependent range, we standardize them within each clip, $\tilde{s}_i = (s_i - \mu_s)/(\sigma_s + \epsilon)$, before converting them into a probability distribution with a softmax of temperature $\tau$,
\begin{equation}
    p_{\text{sal}}(i) = \frac{\exp(\tilde{s}_i / \tau)}{\sum_j \exp(\tilde{s}_j / \tau)}.
\end{equation}
To balance the focus between highly dynamic regions and the static background, we mix the saliency-based probability with a uniform distribution $p_{\text{uni}}(i) = 1/G$, where $G$ is the total number of tokens. The mixing weight $\alpha$ is controlled by a curriculum schedule,
\begin{equation}
    p_{\text{mix}}(i) = \alpha \cdot p_{\text{sal}}(i) + (1 - \alpha) \cdot p_{\text{uni}}(i).
\end{equation}
The curriculum keeps $\alpha$ at $0$ for the first $E_{\text{warmup}}$ epochs, so that masking is initially uniform and the encoder can acquire a generic geometric prior, and then increases it linearly to $\alpha_{\max}$ over the remaining epochs,
\begin{equation}
    \alpha(e) = \alpha_{\max} \cdot \frac{\max(0,\, e - E_{\text{warmup}})}{E_{\text{total}} - E_{\text{warmup}}}, \quad e = 1, \dots, E_{\text{total}}.
\end{equation}
Based on $p_{\text{mix}}$, we generate two complementary masked views. For View A, we sample a subset of tokens $\mathcal{M}_A$ to be masked without replacement according to $p_{\text{mix}}$. For View B, we sample the masked tokens $\mathcal{M}_B$ from the unmasked tokens of View A (\textit{i.e.}, $\mathcal{V}_A = \mathcal{G} \setminus \mathcal{M}_A$) according to the complementary probability $p_{\text{comp}} \propto (1 - p_{\text{mix}})$. The number of masked tokens in View B is controlled by a complement overlap ratio $\gamma \in [0, 1)$, defined as the fraction of $\mathcal{V}_A$ that stays visible in View B as well,
\begin{equation}
    |\mathcal{M}_B| = \max\big(1, \lfloor (1 - \gamma)\, |\mathcal{V}_A| \rfloor \big), \qquad \mathcal{V}_A = \mathcal{G} \setminus \mathcal{M}_A .
\end{equation}
\begin{wrapfigure}{r}{0.48\linewidth}
  \centering
  \includegraphics[width=\linewidth]{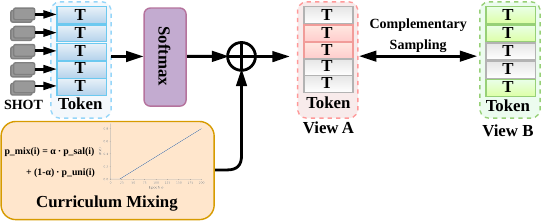}
  \caption{Structure of Curriculum Motion-Saliency Masking (CMSM). Each token's per-frame SHOT descriptors give a motion-saliency score, which a temperature softmax converts into $p_{\text{sal}}$. Curriculum mixing combines $p_{\text{sal}}$ with the uniform distribution $p_{\text{uni}}$ under the schedule weight $\alpha$ to obtain $p_{\text{mix}}$, from which View A is masked directly and View B is masked from View A's visible tokens through complementary sampling.}
  \label{fig,cmsm}
\end{wrapfigure}
Consequently $|\mathcal{V}_A \cap \mathcal{V}_B| = \gamma |\mathcal{V}_A|$ is the set of tokens observed by both views, and $\gamma$ interpolates between exactly complementary visible sets ($\gamma = 0$, where $\mathcal{V}_B = \mathcal{M}_A$) and a View B that masks almost nothing ($\gamma \to 1$).
This construction is deliberately asymmetric. Because $\mathcal{M}_A$ is drawn preferentially from high-saliency tokens while $\mathcal{M}_B$ is drawn from $\mathcal{V}_A$, a token masked in one view tends to remain visible in the other, which is the condition required by CTCP (Section~\ref{sec,ctcp}). Since $\mathcal{M}_B \subseteq \mathcal{V}_A$ by construction, the number of supervised CTCP pairs is deterministically $|\mathcal{M}_A| + (1 - \gamma)(G - |\mathcal{M}_A|)$ at every training step.

\subsection{Asymmetric Encoder-Decoder}
Following the MAE paradigm \citep{he2022masked}, a shared Transformer encoder processes only the visible tokens, $\bm{z}_i = \text{Encoder}(\bm{f}_i + \text{PosEmbed}(\bm{p}_i))$ for $i \in \mathcal{V}$, and a lightweight decoder reconstructs the masked positions, $\bm{d}_i = \text{Decoder}(\bm{z}_{\mathcal{V}} \cup \bm{m}_{\mathcal{M}} + \text{PosEmbed}(\bm{p}_{\mathcal{V} \cup \mathcal{M}}))$. Both views share the same weights. The encoded feature $\bm{z}_i \in \mathbb{R}^{1024}$ of a visible token and the decoded feature $\bm{d}_i \in \mathbb{R}^{512}$ of the same token when masked live in different spaces, and the predictor of Section~\ref{sec,ctcp} bridges them.

\subsection{Reconstruction Targets and Losses}\label{sec,targets}
The decoder is tasked with reconstructing both the appearance and motion information of the masked tokens.

\textbf{Appearance Reconstruction.} We reconstruct the local neighborhood points $\mathcal{N}_i$ of the masked tokens and measure the appearance loss using the Chamfer Distance (CD),
\begin{equation}
    \mathcal{L}_{\text{app}} = \frac{1}{|\mathcal{M}|} \sum_{i \in \mathcal{M}} \text{CD}(\hat{\mathcal{N}}_i, \mathcal{N}_i).
\end{equation}

\textbf{Normal-Flow Motion (NFM) Modeling.} To capture the underlying motion dynamics, we model the local rigid rotation of each token. For a token $i$, we compute the covariance matrix of its neighborhood points at each time step $t$ and perform eigenvalue decomposition to obtain the local reference frame $\bm{R}_t \in \text{SO}(3)$, whose columns are the eigenvectors ordered by decreasing eigenvalue $\lambda_1 \geq \lambda_2 \geq \lambda_3$. Eigenvectors are sign-ambiguous, so we fix the sign of each axis to maximize its inner product with the corresponding axis at $t-1$, and flip the third axis if necessary to enforce $\det(\bm{R}_t) = +1$. This keeps $\bm{R}_t$ a continuous, right-handed function of time. We map $\Delta \bm{R} = \bm{R}_t^\top \bm{R}_{t+1}$ to the Lie algebra $\mathfrak{so}(3)$ to obtain the normal flow vector $\bm{\omega}$,
\begin{equation}
    \bm{\omega} = \log_{\text{SO}(3)}(\Delta \bm{R})^{\vee}.
\end{equation}
Here $\log_{\text{SO}(3)}(\cdot)$ is the matrix logarithm and $(\cdot)^{\vee}: \mathfrak{so}(3) \to \mathbb{R}^3$ extracts the corresponding axis-angle vector.

The decomposition is ill-conditioned whenever two eigenvalues coincide. We therefore require both eigengaps to be well-separated and use
\begin{equation}
    g_i = \frac{\min(\lambda_1 - \lambda_2,\ \lambda_2 - \lambda_3)}{\lambda_1}
\end{equation}
as a conditioning score, excluding any token with $g_i$ below a fixed threshold from $\mathcal{L}_{\text{nfm}}$. Excluded tokens remain supervised by the appearance and TDD losses. We quantify the effect of this exclusion under degraded input in Section~\ref{sec,robustness}.

We regress $\bm{\omega} \in \mathbb{R}^3$ rather than a point-wise displacement field because it is obtained without point-to-point correspondence and is invariant to the number and ordering of points inside a tube. The model predicts $\hat{\bm{\omega}}$, and the NFM loss is
\begin{equation}
    \mathcal{L}_{\text{nfm}} = \frac{1}{|\mathcal{M}|} \sum_{i \in \mathcal{M}} \text{SmoothL1}(\hat{\bm{\omega}}_i, \bm{\omega}_i).
\end{equation}
We adopt SmoothL1 on $\bm{\omega}$ rather than the geodesic distance on SO(3), whose gradient vanishes near a rotation of $0$ and becomes singular near $\pi$.

\textbf{Temporal Descriptor Difference (TDD).} As a complementary motion target we predict the temporal difference of the SHOT descriptors, $\Delta \bm{h}_i = \bm{h}_{i,t+1} - \bm{h}_{i,t}$,
\begin{equation}
    \mathcal{L}_{\text{tdd}} = \frac{1}{|\mathcal{M}|} \sum_{i \in \mathcal{M}} \text{SmoothL1}(\Delta \hat{\bm{h}}_i, \Delta \bm{h}_i).
\end{equation}
TDD summarizes how much the local histogram changed, and is therefore sensitive to deformation and occlusion as well as density change, but it does not recover the axis and angle of the motion. NFM supplies that structure, and conversely remains undefined on tokens excluded by $g_i$, where TDD is still informative. The total motion loss is $\mathcal{L}_{\text{motion}} = \mathcal{L}_{\text{nfm}} + \mathcal{L}_{\text{tdd}}$.

\subsection{Cross-view Token Consistency Prediction (CTCP)}\label{sec,ctcp}
To further exploit the complementary nature of View A and View B, we introduce the CTCP module. The intuition is that the reconstructed features of a masked token in one view should be consistent with the encoded features of the same token when it is visible in the other view.

Specifically, let $\bm{d}_i^A \in \mathbb{R}^{C_d}$ be the decoded feature of a masked token $i \in \mathcal{M}_A$ in View A. If this token is visible in View B, we have its encoded feature $\bm{z}_i^B \in \mathbb{R}^{C_e}$. The eligible set for this direction is therefore
\begin{equation}
    \mathcal{C}_{A \to B} = \mathcal{M}_A \cap \mathcal{V}_B ,
\end{equation}
which by the construction of Section~\ref{sec,cmsm} equals $\mathcal{M}_A$ and is never empty. A predictor network maps the decoded feature into the encoder feature space, $\hat{\bm{z}}_i^B = \text{Predictor}(\bm{d}_i^A) \in \mathbb{R}^{C_e}$, absorbing both the dimensionality mismatch ($C_d = 512$ versus $C_e = 1024$) and the representational gap between decoder and encoder. The CTCP loss averages the negative cosine similarity between the predicted and target features over the eligible set,
\begin{equation}
    \mathcal{L}_{\text{ctcp}}^{A \to B} = \frac{1}{|\mathcal{C}_{A \to B}|} \sum_{i \in \mathcal{C}_{A \to B}} \left( 1 - \frac{\hat{\bm{z}}_i^B \cdot \text{sg}(\bm{z}_i^B)}{\|\hat{\bm{z}}_i^B\|_2 \, \|\text{sg}(\bm{z}_i^B)\|_2} \right),
\end{equation}
where $\text{sg}(\cdot)$ denotes the stop-gradient operation. The symmetric loss $\mathcal{L}_{\text{ctcp}}^{B \to A}$ is defined identically over $\mathcal{C}_{B \to A} = \mathcal{M}_B$, and the total CTCP loss is $\mathcal{L}_{\text{ctcp}} = \frac{1}{2} (\mathcal{L}_{\text{ctcp}}^{A \to B} + \mathcal{L}_{\text{ctcp}}^{B \to A})$.

Following non-contrastive siamese learning \citep{grill2020byol, chen2021simsiam}, three design choices prevent collapse: a stop-gradient on the target branch, an explicit predictor that absorbs the encoder--decoder mismatch, and the asymmetric complementary sampling of Section~\ref{sec,cmsm}, which keeps the number of CTCP pairs fixed. Table~\ref{tab,ablation_ctcp} confirms that removing the predictor causes a collapse-like degradation, whereas replacing the stop-gradient target with an EMA encoder is a viable but slightly weaker alternative.

We operate CTCP at the token level rather than pooling to a clip-level descriptor, because temporal action segmentation and semantic segmentation require localized discriminative structure.

\subsection{Overall Objective}
The overall self-supervised training objective of MoSaiC is a weighted sum of the appearance, motion, and CTCP losses from both views,
\begin{equation}
    \mathcal{L} = \frac{1}{2} \sum_{v \in \{A, B\}} (\mathcal{L}_{\text{app}}^v + \mathcal{L}_{\text{motion}}^v) + \lambda_{\text{ctcp}} \mathcal{L}_{\text{ctcp}},
\end{equation}
where $\lambda_{\text{ctcp}}$ is a hyperparameter balancing the consistency loss.

\section{Experiments}\label{sec,experiments}
\subsection{Datasets and Evaluation Metrics}
We evaluate MoSaiC on three downstream tasks across four datasets, action recognition on MSR-Action3D \citep{li2010action}, SHREC'17 \citep{desmedt2017shrec}, and NvGesture \citep{molchanov2016online}, temporal action segmentation on HOI4D \citep{liu2022hoi4d}, and semantic segmentation on HOI4D.

\textbf{MSR-Action3D} contains 567 point cloud videos of 20 action categories performed by 10 subjects. We follow the standard cross-subject evaluation protocol, using subjects 1, 3, 5, 7, 9 for training and the rest for testing. The evaluation metric is Top-1 accuracy.

\textbf{SHREC'17} is a dynamic hand-gesture dataset containing 2,800 depth and skeletal sequences of 14 gestures performed by 28 subjects. We use its depth sequences as point cloud videos and follow the official recognition protocol. The evaluation metric is Top-1 accuracy.

\textbf{NvGesture} is a multimodal dynamic hand-gesture dataset acquired with depth, color, and stereo-IR sensors. It contains 25 gesture categories performed by 20 subjects. We use the depth modality converted to point cloud videos and follow the official subject-independent split. The evaluation metric is Top-1 accuracy.

\textbf{HOI4D} is a large-scale 4D ego-centric dataset for human-object interaction. For \textbf{temporal action segmentation}, the dataset provides frame-level annotations for 19 action classes. We report Frame-wise Accuracy (Frame-Acc), Segmental Edit Distance (Edit), and Segmental F1 scores at overlapping thresholds of 10\%, 25\%, and 50\% (F1@\{10, 25, 50\}), following the per-class breakdown we use in Section~\ref{sec,experiments}, F1@$k$ is computed per action class and then averaged over the 19 classes. For \textbf{semantic segmentation}, the dataset annotates 38 indoor semantic categories in total, following the official 4D point cloud video segmentation setting, we evaluate on the 14 selected categories (7 object and 7 background). We report the mean Intersection over Union (mIoU).

\subsection{Implementation Details}
\textbf{Network Architecture.} The tube embedding module uses a spatial radius $r = 0.1$, a maximum of $k = 32$ neighbors, a spatial stride of $32$, a temporal kernel size of $3$, and a temporal stride of $2$ (except for semantic segmentation where the temporal stride is $1$). The encoder is a 5-layer Transformer with an embedding dimension of $1024$, 8 heads, and an MLP dimension of $2048$, giving 43.6M parameters in total including the tube embedding. The decoder is a 4-layer Transformer with an embedding dimension of $512$ and is discarded after pre-training.

\textbf{Pre-training.} The model is pre-trained using the AdamW optimizer with a base learning rate of $1\times 10^{-3}$ and a weight decay of $0.05$. We use a cosine learning rate schedule with a 10-epoch linear warmup for a total of 200 epochs. The base mask ratio is set to $0.6$ on all datasets. For CMSM, the temperature $\tau$ is $1.0$, the maximum curriculum weight $\alpha_{\max}$ is $0.8$, and the curriculum warmup is $20$ epochs. The complement overlap ratio $\gamma$ is $0.1$. The CTCP loss weight $\lambda_{\text{ctcp}}$ is $0.25$.

\textbf{Fine-tuning.} For downstream tasks, we fine-tune the pre-trained encoder along with a task-specific head. We use the AdamW optimizer with a learning rate of $5\times 10^{-4}$ and a cosine schedule. The encoder learning rate is scaled down by a factor (\textit{e.g.}, $2\times 10^{-4}$). Label smoothing is applied for classification losses.

\subsection{Comparison with State-of-the-Art}

\textbf{Action Recognition.} Table~\ref{tab,action_recognition} compares MoSaiC with representative supervised backbones and self-supervised pre-training methods. On the two hand-gesture benchmarks, MoSaiC attains the best accuracy, reaching 91.9\% on SHREC'17 and 89.3\% on NvGesture. This improves over the from-scratch P4Transformer backbone by 2.8 and 3.2 points respectively, and over the strongest self-supervised competitor, C2P, by 0.7 and 0.5 points. Hand gestures are dominated by fine-grained finger articulation against a largely static hand silhouette, so these benchmarks reward precisely the property that CMSM is designed to induce, concentrating the pretext task on the small subset of tokens that actually move, rather than spending capacity on background geometry that is trivially reconstructible.

On MSR-Action3D, MoSaiC reaches 95.18\%, 4.24 points above the from-scratch backbone and, notably, also ahead of the strongest masked-modeling baseline, C2P, though by a narrower margin (+0.42) than on the two hand-gesture benchmarks, we return to why this particular benchmark yields the smallest relative margin among the three action-recognition datasets in Section~\ref{sec,discussion}. The gap widens substantially on the dense spatio-temporal tasks reported next, where temporal reasoning is indispensable.

\begin{table*}[ht]
\caption{Action recognition results (Top-1 accuracy, \%) on MSR-Action3D, SHREC'17, and NvGesture. All self-supervised methods, including MoSaiC, use the P4Transformer backbone, whose supervised from-scratch result serves as the common no-pre-training reference.}\label{tab,action_recognition}
\begin{NiceTabular*}{\textwidth}{@{}LCCC@{}}
\toprule
Method & MSR-Action3D & SHREC'17 & NvGesture \\
\midrule
\multicolumn{4}{@{}l}{\textit{Supervised backbones (trained from scratch)}} \\
PSTNet \cite{fan2021pstnet} & 91.20 & 88.5 & 85.9 \\
P4Transformer \cite{fan2021p4transformer} & 90.94 & 89.1 & 86.1 \\
PST-Transformer \cite{fan2022pst} & 93.73 & 90.3 & 87.0 \\
Kinet \cite{zhong2022kinet} & 93.27 & 90.1 & 86.8 \\
\midrule
\multicolumn{4}{@{}l}{\textit{Self-supervised pre-training (P4Transformer backbone)}} \\
PointCMP \cite{shen2023pointcmp} & 93.27 & 90.6 & 88.5 \\
MaST-Pre \cite{shen2023mastpre} & 94.08 & 90.8 & 88.1 \\
C2P \cite{zhang2023c2p} & 94.76 & 91.2 & 88.8 \\
\rowcolor{mosaicrow}
\textbf{MoSaiC (Ours)} & \textbf{95.18} & \textbf{91.9} & \textbf{89.3} \\
\bottomrule
\end{NiceTabular*}
\end{table*}

\textbf{Temporal Action Segmentation.} Because pre-training methods for point cloud videos are built on different backbones, Table~\ref{tab,action_segmentation} groups the HOI4D action segmentation results by backbone, and additionally instantiates MoSaiC on both backbones so that every group ends with a directly comparable pre-training result. On its default P4Transformer backbone, MoSaiC reaches 81.9\% Frame-Acc, 83.6 Edit, and 86.2/83.4/76.8 on F1@\{10, 25, 50\}, outperforming the strongest backbone-matched baseline, M2PSC, by 6.0 Frame-Acc and 10.9 F1@50. To verify that this gain is not an artifact of a particular backbone, we also instantiate MoSaiC on PPTr, a backbone with a longer effective temporal receptive field. There, MoSaiC still improves over the strongest PPTr-based baseline, C2P, by 2.4 Frame-Acc and 3.3 F1@50, confirming that the pretext task contributes on top of a stronger encoder as well, albeit with a smaller margin than on P4Transformer. Relative to the from-scratch P4Transformer backbone, the most informative pattern is that the margin grows monotonically with the overlap threshold, MoSaiC leads by 12.4 points at F1@10 but by 18.6 points at F1@50. A gain concentrated at the strict threshold cannot be explained by simply labelling more frames correctly, it requires predicting segments whose temporal extents align tightly with the ground-truth boundaries.

This is exactly the behaviour that NFM is designed to produce. By supervising the relative rotation of each token's local reference frame in $\mathfrak{so}(3)$, the pretext task forces the encoder to represent \emph{when} a local part changes its motion state, which is the cue that delimits an action boundary. An unstructured change-magnitude target, such as MaST-Pre's cardinality difference or our own TDD, registers that something changed but not how the local frame reoriented, and therefore leaves boundary transitions between two motion states of similar magnitude largely unsupervised. The comparatively modest scores of MaST-Pre and M2PSC under the strict F1@50 metric (63.8 and 65.9, both on the same P4Transformer backbone as MoSaiC) are consistent with this interpretation.

\begin{table*}[ht]
\caption{Temporal action segmentation results on HOI4D. Methods are grouped by backbone, and MoSaiC is instantiated on both PPTr and P4Transformer (its default backbone) to verify that its gains generalize across backbones. Baseline numbers for STRL, VideoMAE, C2P, and M2PSC are as reported in their respective papers.}\label{tab,action_segmentation}
\begin{NiceTabular*}{\textwidth}{@{}LCCCCC@{}}
\toprule
Method & Frame-Acc (\%) & Edit & F1@10 & F1@25 & F1@50 \\
\midrule
PPTr \cite{wen2022pptr} (scratch) & 77.4 & 80.1 & 81.7 & 78.5 & 69.5 \\
+ STRL \cite{huang2021strl} & 78.4 & 79.1 & 81.8 & 78.6 & 69.7 \\
+ VideoMAE \cite{tong2022videomae} & 78.6 & 80.2 & 81.9 & 78.7 & 69.9 \\
+ C2P \cite{zhang2023c2p} & 81.1 & 84.0 & 85.4 & 82.5 & 74.1 \\
\rowcolor{mosaicrow}
\textbf{+ MoSaiC (Ours)} & \textbf{83.5} & \textbf{84.9} & \textbf{88.1} & \textbf{84.1} & \textbf{77.4} \\
\midrule
P4Transformer \cite{fan2021p4transformer} (scratch) & 71.2 & 73.1 & 73.8 & 69.2 & 58.2 \\
+ C2P \cite{zhang2023c2p} & 73.5 & 76.8 & 77.2 & 72.9 & 62.4 \\
+ MaST-Pre \cite{shen2023mastpre} & 74.1 & 75.4 & 76.6 & 73.4 & 63.8 \\
+ M2PSC \cite{han2024m2psc} & 75.9 & 77.1 & 78.4 & 74.1 & 65.9 \\
\rowcolor{mosaicrow}
\textbf{+ MoSaiC (Ours)} & \textbf{81.9} & \textbf{83.6} & \textbf{86.2} & \textbf{83.4} & \textbf{76.8} \\
\bottomrule
\end{NiceTabular*}
\end{table*}

\textbf{Where the Action Segmentation Gain Comes From.} The aggregate 13.0-point F1@50 margin over MaST-Pre in Table~\ref{tab,action_segmentation} could in principle come from a uniform improvement across all 19 HOI4D action classes, or from a concentrated improvement on a subset of them, these two scenarios support very different explanations of why MoSaiC helps. Table~\ref{tab,perclass} breaks down the per-class F1@50 for the four classes with the largest gain and the three with the smallest gain (the remaining 12 classes fall between these extremes and average $+13.2$, close to the overall margin). The gain is clearly concentrated, the four largest-gain classes, \textit{Turnon}, \textit{Papercut}, \textit{Dump}, and \textit{Open}, all involve a sustained rotation, either of a hand-held object about the wrist, as in tipping a bucket or kettle, or of an articulated part about its joint axis, as in turning a knob, closing the blades of a pair of scissors, or swinging a lid or a laptop screen, these classes improve by 16.9 to 22.4 points. The three smallest-gain classes, \textit{Carry}, \textit{Reachout}, and \textit{Rest}, improve by only 2.1 to 4.8 points, each is characterised either by translation-dominated hand motion with a nearly constant object orientation or by the absence of motion altogether, so the orientation-change information that NFM supplies adds little beyond what appearance and TDD cues already provide.

This pattern is precisely what we would expect if the aggregate gain is driven by NFM rather than by a generic effect of more pre-training or more parameters, classes whose boundaries are defined by rotation dynamics benefit most from a pretext task that explicitly supervises rotation, while classes that are translation-dominated or static benefit comparatively little. It also clarifies a point left implicit in the discussion of Table~\ref{tab,action_segmentation}, namely why the margin over MaST-Pre grows with the overlap threshold, rotation-heavy classes require precise boundary localization to be counted as correct at F1@50, so a method that captures orientation dynamics should disproportionately gain at the strict threshold, which is exactly the trend observed.

\begin{table*}[ht]
\caption{Per-class F1@50 on HOI4D temporal action segmentation, showing the four classes with the largest MoSaiC gain over MaST-Pre and the three with the smallest. The overall row is the class-wise mean reported in Table~\ref{tab,action_segmentation}, the 12 classes omitted here have gains between $+4.8$ and $+16.9$ and average $+13.2$.}\label{tab,perclass}
\begin{NiceTabular*}{\textwidth}{@{}LCCC@{}}
\toprule
Action Class & MaST-Pre & MoSaiC & $\Delta$ \\
\midrule
Turnon & 41.2 & 63.6 & +22.4 \\
Papercut & 47.5 & 67.6 & +20.1 \\
Dump & 52.1 & 70.7 & +18.6 \\
Open & 58.3 & 75.2 & +16.9 \\
\midrule
Carry & 74.6 & 79.4 & +4.8 \\
Reachout & 73.1 & 76.5 & +3.4 \\
Rest & 81.9 & 84.0 & +2.1 \\
\midrule
\rowcolor{mosaicrow}
\textbf{All 19 classes (mean)} & \textbf{63.8} & \textbf{76.8} & \textbf{+13.0} \\
\bottomrule
\end{NiceTabular*}
\end{table*}

\begin{figure*}[t]
  \centering
  \includegraphics[width=\textwidth]{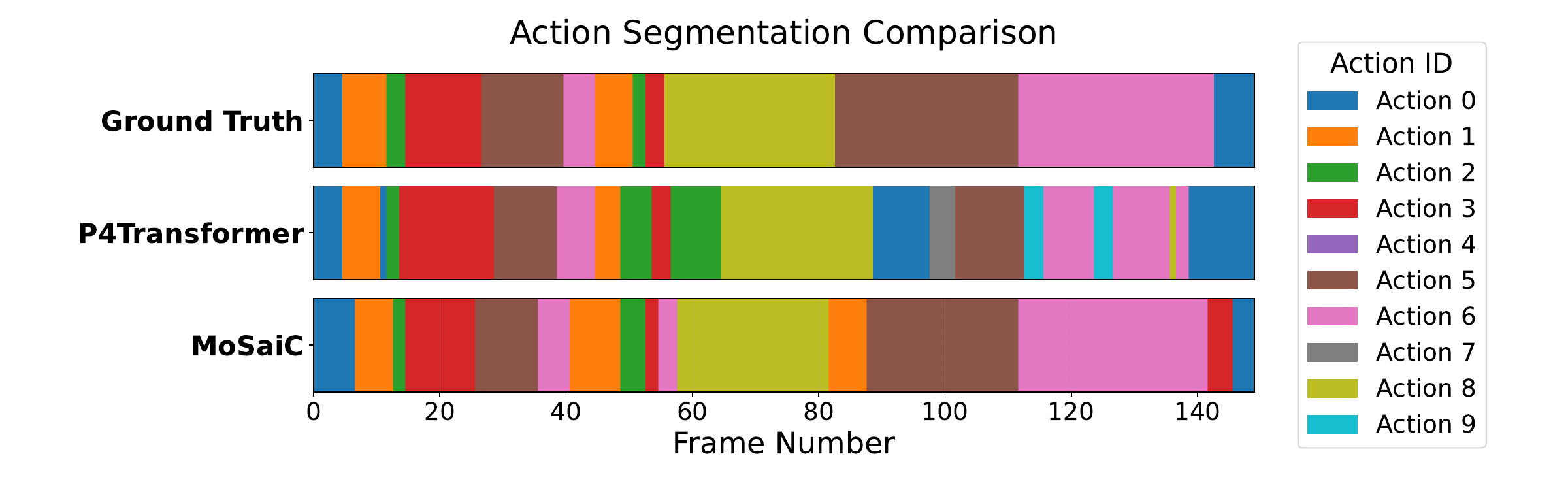}
  \caption{Qualitative comparison of frame-level action segmentation on a representative $\sim$150-frame HOI4D clip. From top to bottom, Ground Truth, the P4Transformer baseline, and MoSaiC. The horizontal axis is the frame index, and each color denotes a distinct action class, so a correct prediction requires both the color-block boundaries to align with those of Ground Truth (accurate temporal localization) and the color identity to match (correct class prediction).}
  \label{fig,action_comparison}
\end{figure*}

Figure~\ref{fig,action_comparison} complements this per-class quantitative breakdown with a qualitative view. The P4Transformer baseline tends to oversegment, splitting several ground-truth segments into shorter, inconsistently colored fragments, and misassigns the class of some segments altogether. MoSaiC's predicted color bands track Ground Truth much more closely, both in where segment boundaries fall and in which class label each segment receives, visually corroborating the more precise motion-boundary localization that we attribute to NFM above.

\textbf{Semantic Segmentation.} As with action segmentation, Table~\ref{tab,semantic_segmentation} groups the HOI4D semantic segmentation results by backbone. On its default P4Transformer backbone, MoSaiC achieves 43.7\% mIoU, exceeding its strongest backbone-matched competitor, M2PSC, by 1.4 mIoU, and MaST-Pre by 3.4 mIoU. Instantiated on the PPTr backbone, MoSaiC reaches 44.3\% mIoU, exceeding the strongest PPTr-based baseline, C2P, by 2.0 mIoU, confirming that the benefit of the pretext task is not confined to a single backbone. This result deserves emphasis because semantic segmentation is a static per-point labelling task in which motion is not directly required at inference time. A motion-guided pretext task could plausibly bias the encoder away from the fine geometric detail that per-point labelling needs, yet we observe the opposite.

We attribute this to the geometric formulation of our motion target. Because NFM is computed from a local PCA reference frame, predicting it requires the encoder to retain an accurate description of the local surface orientation of each token, not merely a coarse notion of displacement. Surface orientation is simultaneously the primary cue for distinguishing object parts, so the motion objective and the semantic objective reinforce rather than compete with each other. The fact that MaST-Pre improves only marginally over its backbone on this task (40.3 versus 40.1 mIoU) supports the view that the benefit comes from the geometric nature of the target rather than from motion supervision in general.

\begin{table}[ht]
\caption{Semantic segmentation results (mIoU, \%) on HOI4D. Methods are grouped by backbone, and MoSaiC is instantiated on both PPTr and P4Transformer (its default backbone) to verify that its gains generalize across backbones. Baseline numbers for STRL, VideoMAE, C2P, and M2PSC are as reported in their respective papers.}\label{tab,semantic_segmentation}
\begin{NiceTabular*}{\textwidth}{@{}LC@{}}
\toprule
Method & mIoU (\%) \\
\midrule
PPTr \cite{wen2022pptr} (scratch) & 41.0 \\
+ STRL \cite{huang2021strl} & 41.2 \\
+ VideoMAE \cite{tong2022videomae} & 41.3 \\
+ C2P \cite{zhang2023c2p} & 42.3 \\
\rowcolor{mosaicrow}
\textbf{+ MoSaiC (Ours)} & \textbf{44.3} \\
\midrule
P4Transformer \cite{fan2021p4transformer} (scratch) & 40.1 \\
+ MaST-Pre \cite{shen2023mastpre} & 40.3 \\
+ C2P \cite{zhang2023c2p} & 41.4 \\
+ M2PSC \cite{han2024m2psc} & 42.3 \\
\rowcolor{mosaicrow}
\textbf{+ MoSaiC (Ours)} & \textbf{43.7} \\
\bottomrule
\end{NiceTabular*}
\end{table}

\subsection{Label Efficiency and Linear Probing}
A key motivation of self-supervised pre-training is to reduce the dependence on costly 4D annotations. We therefore evaluate MoSaiC under two additional protocols. \textbf{(i) Label efficiency,} we fine-tune the pre-trained encoder using only a fraction (10\% and 50\%) of the labeled training data, and compare against training from scratch and against representative self-supervised counterparts. \textbf{(ii) Linear probing,} we freeze the pre-trained encoder and train only a linear classifier on top of it, which directly measures the linear separability and quality of the learned representations. Table~\ref{tab,label_efficiency} reports the results on MSR-Action3D action recognition. With only 10\% of the labels, MoSaiC attains 77.8\% accuracy against 62.4\% for the from-scratch baseline, a margin of 15.4 points that is more than three times the 4.24-point margin observed under full supervision. The advantage narrows to 8.3 points at 50\% labels, showing that the value of pre-training is largest exactly where annotation is scarcest. MoSaiC also leads the strongest self-supervised counterpart by 3.2 points in the 10\% regime, indicating that the gain is not merely an artifact of pre-training in general but of the specific pretext design.

Linear probing isolates this further. Because the encoder is frozen, the resulting 86.9\% accuracy measures how much action-discriminative structure is already linearly accessible in the representation before any task-specific adaptation. MoSaiC exceeds MaST-Pre by 2.6 points and PointCMP by 3.3 points under this protocol. Since the three methods share the same backbone and the same 200-epoch pre-training schedule, the difference is attributable to the objectives themselves, cross-view token consistency supplies an instance-discriminative signal that pure reconstruction lacks, and it operates at token granularity, so the resulting structure survives the pooling step used by the linear classifier.

\begin{table*}[ht]
\caption{Label-efficiency and linear-probing results for action recognition on MSR-Action3D (Top-1 accuracy, \%). FT denotes fine-tuning with the indicated fraction of labels. LP denotes linear probing. The FT (100\%) column repeats the fully-supervised results of Table~\ref{tab,action_recognition}.}\label{tab,label_efficiency}
\begin{NiceTabular*}{\textwidth}{@{}LCCCC@{}}
\toprule
Method & FT (10\%) & FT (50\%) & FT (100\%) & LP (100\%) \\
\midrule
Scratch (no pre-training) & 62.4 & 82.1 & 90.94 & N/A \\
MaST-Pre \cite{shen2023mastpre} & 74.6 & 88.7 & 94.08 & 84.3 \\
PointCMP \cite{shen2023pointcmp} & 73.9 & 88.2 & 93.27 & 83.6 \\
\rowcolor{mosaicrow}
\textbf{MoSaiC (Ours)} & \textbf{77.8} & \textbf{90.4} & \textbf{95.18} & \textbf{86.9} \\
\bottomrule
\end{NiceTabular*}
\end{table*}

\subsection{Transfer Learning}
\textbf{Cross-dataset transfer.} To assess generalization beyond the pre-training domain, we transfer an encoder pre-trained on HOI4D to the SHREC'17 and NvGesture hand-gesture benchmarks, and fine-tune it using their labeled training sets. Table~\ref{tab,transfer} reports these transfer-learning results. MoSaiC reaches 91.4\% on SHREC'17 and 87.9\% on NvGesture, improving over the from-scratch baseline by 2.3 and 1.8 points and over MaST-Pre by 1.3 and 1.2 points. This setting is a demanding test of generality, because HOI4D consists of egocentric human-object interactions at room scale whereas the two targets consist of close-range hand gestures, the two domains share neither object categories, spatial scale, nor capture viewpoint. That the pre-trained encoder still transfers positively indicates that what MoSaiC learns is a generic prior over how local geometry deforms over time, rather than a memorized inventory of HOI4D-specific shapes. Such a prior remains valid under a change of domain, which is consistent with the fact that the transfer margins are of the same order as the in-domain margins rather than collapsing.

\begin{table*}[ht]
\caption{Cross-dataset transfer for action recognition (Top-1 accuracy, \%). Models are pre-trained on the source and fine-tuned on each target.}\label{tab,transfer}
\begin{NiceTabular*}{\textwidth}{@{}LCC@{}}
\toprule
Method & Source $\to$ Target & Acc (\%) \\
\midrule
Scratch (no pre-training) & N/A $\to$ SHREC'17 / NvGesture & 89.1 / 86.1 \\
MaST-Pre \cite{shen2023mastpre} & HOI4D $\to$ SHREC'17 / NvGesture & 90.1 / 86.7 \\
\rowcolor{mosaicrow}
\textbf{MoSaiC (Ours)} & HOI4D $\to$ SHREC'17 / NvGesture & \textbf{91.4 / 87.9} \\
\bottomrule
\end{NiceTabular*}
\end{table*}

\subsection{Efficiency Analysis}
Table~\ref{tab,efficiency} compares model complexity and pre-training cost with representative self-supervised methods. Because MoSaiC discards the decoder, the predictor, and all motion-target computations after pre-training, its fine-tuning and inference cost is identical to the underlying backbone, all three methods in Table~\ref{tab,efficiency} share the same 43.6M encoder parameters and 78.4G inference FLOPs.

The only overhead is therefore confined to pre-training, where MoSaiC requires 30.2 hours against 21.5 hours for MaST-Pre ($1.4\times$), almost entirely because two complementary views are encoded instead of one, the saliency scores and NFM targets reuse quantities (SHOT descriptors, local PCA statistics) that the tokenizer already computes for the single-view baselines, so they add negligible cost on top of the dual-view forward pass.

\begin{table}[ht]
\caption{Efficiency comparison. Params (M) and FLOPs (G) are measured on the encoder used for downstream inference, Pre-train is the wall-clock pre-training time in hours.}\label{tab,efficiency}
\begin{NiceTabular*}{\textwidth}{@{}LCCC@{}}
\toprule
Method & Params & FLOPs & Pre-train \\
\midrule
MaST-Pre \cite{shen2023mastpre} & 43.6 & 78.4 & 21.5 \\
PointCMP \cite{shen2023pointcmp} & 43.6 & 78.4 & 19.8 \\
\rowcolor{mosaicrow}
\textbf{MoSaiC (Ours)} & \textbf{43.6} & \textbf{78.4} & \textbf{30.2} \\
\bottomrule
\end{NiceTabular*}
\end{table}

\subsection{Ablation Studies}
We conduct ablation studies to analyze the contribution of each component in MoSaiC. Unless otherwise specified, action recognition is evaluated on MSR-Action3D (Top-1 accuracy) and the dense prediction tasks on HOI4D (action segmentation F1@10 and semantic segmentation mIoU).

\textbf{Component Analysis.} Table~\ref{tab,ablation_components} reports a cumulative ablation that progressively adds each proposed component to the baseline. The baseline adopts random masking with appearance (Chamfer) reconstruction plus the TDD target, which plays the role of the change-magnitude target used by prior work, so that the NFM column isolates the effect of our own motion target rather than of motion supervision in general. Adding Curriculum Motion-Saliency Masking (CMSM) directs the pretext task toward dynamic regions, further adding Normal-Flow Motion (NFM) modeling injects explicit geometric motion supervision, and finally Cross-view Token Consistency Prediction (CTCP) enforces cross-view consistency. Each component brings consistent improvements across all three tasks, and the full MoSaiC achieves the best results, validating that the three components are complementary.

The magnitude of each increment is informative. CMSM alone contributes the single largest gain on action segmentation ($78.6 \rightarrow 82.1$ F1@10, $+3.5$), which confirms that \emph{where} the model is asked to reconstruct matters as much as \emph{what} it is asked to reconstruct, redirecting the same reconstruction budget toward motion-salient tokens improves temporal understanding without adding any new loss term. NFM adds a further $+2.6$ F1@10 and $+1.1$ mIoU, and is the only component that improves the two dense tasks by comparable amounts, consistent with its target being geometric rather than purely temporal. CTCP contributes the smallest increment on the dense tasks ($+1.5$ F1@10) but the largest single increment on recognition ($+2.0$), matching its role as an instance-level discriminative signal rather than a localization signal. The dense-task increments do shrink as components accumulate ($+3.5$, $+2.6$, $+1.5$ F1@10), as expected once the earlier components have already absorbed the easiest headroom, but the ordering across metrics is not the same for the three components, CMSM is strongest on action segmentation, NFM is the only component that lifts the two dense tasks by comparable amounts, and CTCP is strongest on recognition. This dissociation, rather than the raw size of the increments, is what indicates that the three objectives supervise largely non-overlapping aspects of the representation.

\begin{table*}[ht]
\caption{Cumulative component ablation. We progressively add each proposed component to the baseline (random masking + appearance reconstruction + TDD). Action Rec., MSR-Action3D Top-1 accuracy, Action Seg., HOI4D F1@10, Sem. Seg., HOI4D mIoU.}\label{tab,ablation_components}
\begin{NiceTabular*}{\textwidth}{@{}LcccCCC@{}}
\toprule
Configuration & CMSM & NFM & CTCP & Action Rec. & Action Seg. & Sem. Seg. \\
\midrule
Baseline & & & & 91.3 & 78.6 & 40.4 \\
+ CMSM & \checkmark & & & 92.4 & 82.1 & 41.8 \\
+ CMSM + NFM & \checkmark & \checkmark & & 93.2 & 84.7 & 42.9 \\
\rowcolor{mosaicrow}
\textbf{MoSaiC (full)} & \checkmark & \checkmark & \checkmark & \textbf{95.18} & \textbf{86.2} & \textbf{43.7} \\
\bottomrule
\end{NiceTabular*}
\end{table*}

\textbf{Masking Strategy.} Table~\ref{tab,ablation_masking} compares different masking strategies. Replacing random masking with pure motion-saliency masking already yields $+1.2$ F1@10, and adding the curriculum schedule contributes a further $+1.1$, for a total of $+2.3$ over the random baseline. The second increment is the one that validates our design choice. Motion-saliency masking without a curriculum aggressively hides the most dynamic tokens from the very first epoch, at which point the encoder has not yet learned a usable geometric prior and the reconstruction target is close to unpredictable, the resulting gradients are dominated by noise. By annealing $\alpha$ from $0$ to $\alpha_{\max}$, CMSM lets the model first acquire generic structure under an essentially uniform masking distribution and only then shifts the difficulty toward dynamic regions. The consistent ordering across both dense tasks indicates that this is a genuine optimization effect rather than a task-specific artifact.

\begin{table}[ht]
\caption{Ablation on masking strategies. Action Seg., HOI4D F1@10, Sem. Seg., HOI4D mIoU.}\label{tab,ablation_masking}
\begin{NiceTabular*}{\textwidth}{@{}LCC@{}}
\toprule
Masking Strategy & Action Seg. & Sem. Seg. \\
\midrule
Random & 83.9 & 42.1 \\
Motion Saliency (MS) & 85.1 & 42.8 \\
\rowcolor{mosaicrow}
\textbf{MS + Curriculum (CMSM)} & \textbf{86.2} & \textbf{43.7} \\
\bottomrule
\end{NiceTabular*}
\end{table}

\textbf{Motion Targets.} Table~\ref{tab,ablation_motion} evaluates the impact of different motion reconstruction targets. Any form of motion supervision improves over appearance-only reconstruction, but the two targets are not interchangeable, NFM alone outperforms TDD alone by 1.1 F1@10 and 0.7 mIoU. This gap supports our central claim that an unstructured change-magnitude target is too coarse. TDD records how much each histogram bin of a token changed, but two motions with the same descriptor perturbation, for instance a rotation about one axis and a rotation of similar magnitude about another, are assigned indistinguishable targets, whereas NFM assigns each a distinct element of $\mathfrak{so}(3)$ with an explicit axis and angle.

Combining both targets is nonetheless better than either alone ($+0.8$ F1@10 over NFM), which indicates that they are complementary rather than competing. The two cover disjoint failure modes, TDD remains informative under occlusion, dis-occlusion, and deformation, precisely the conditions in which the eigengap criterion $g_i$ of Section~\ref{sec,targets} withdraws NFM supervision because the local frame is ill-conditioned, conversely NFM resolves the direction of a rigid motion that TDD can only report as a magnitude. Supervising both therefore yields motion signals over a broader range of conditions than either target can cover in isolation.

\begin{table}[ht]
\caption{Ablation on motion reconstruction targets. Action Seg., HOI4D F1@10, Sem. Seg., HOI4D mIoU.}\label{tab,ablation_motion}
\begin{NiceTabular*}{\textwidth}{@{}LCC@{}}
\toprule
Motion Target & Action Seg. & Sem. Seg. \\
\midrule
None (Appearance only) & 82.4 & 41.5 \\
TDD & 84.3 & 42.4 \\
NFM & 85.4 & 43.1 \\
\rowcolor{mosaicrow}
\textbf{TDD + NFM} & \textbf{86.2} & \textbf{43.7} \\
\bottomrule
\end{NiceTabular*}
\end{table}

\textbf{CTCP Design.} The preceding ablations validate CMSM and NFM but leave CTCP, our third component, supported only by its contribution to Table~\ref{tab,ablation_components}. Table~\ref{tab,ablation_ctcp} isolates the design choices identified in Section~\ref{sec,ctcp}. The row without CTCP reproduces the \mbox{CMSM + NFM} configuration of Table~\ref{tab,ablation_components}. Removing the predictor while keeping the stop-gradient target collapses performance far below even that configuration (78.1 F1@10, 39.1 mIoU, versus 84.7/42.9 without CTCP at all), confirming that an explicit predictor is not an optional refinement but a necessary condition for the objective to avoid the degenerate solution discussed above, without it, the shared pressure toward a trivial constant representation actively harms the features learned by the appearance and motion losses rather than merely failing to help. Replacing the stop-gradient online target with an EMA target is a viable alternative (86.0/43.5) but is marginally worse and adds a second set of momentum-updated weights, so we retain the simpler stop-gradient design. The complement overlap ratio $\gamma$ has a clear optimum at $0.1$, and the two directions in which it degrades have different causes. At $\gamma = 0$ the number of CTCP pairs is maximal, but View B then observes exactly $\mathcal{M}_A$, \textit{i.e.}, only the high-saliency tokens, and loses the static context that anchors its own reconstruction losses, since those losses are what keep the CTCP target branch from drifting, the targets themselves become less reliable. At $\gamma \geq 0.3$ the opposite happens, View B masks progressively fewer tokens, its reconstruction task becomes easy, and the number of $B \to A$ CTCP pairs falls by $(1-\gamma)(G - |\mathcal{M}_A|)$.

\begin{table*}[ht]
\caption{Ablation on CTCP design. Action Rec., MSR-Action3D Top-1 accuracy, Action Seg., HOI4D F1@10, Sem. Seg., HOI4D mIoU.}\label{tab,ablation_ctcp}
\begin{NiceTabular*}{\textwidth}{@{}LCCC@{}}
\toprule
Configuration & Action Rec. & Action Seg. & Sem. Seg. \\
\midrule
w/o CTCP (dual view only) & 93.2 & 84.7 & 42.9 \\
w/o predictor & 90.9 & 78.1 & 39.1 \\
EMA target instead of stop-gradient & 94.9 & 86.0 & 43.5 \\
$\gamma = 0$ & 94.6 & 85.0 & 43.0 \\
$\gamma = 0.3$ & 94.9 & 85.8 & 43.4 \\
$\gamma = 0.5$ & 94.7 & 85.3 & 43.2 \\
\rowcolor{mosaicrow}
\textbf{$\gamma = 0.1$ (MoSaiC)} & \textbf{95.18} & \textbf{86.2} & \textbf{43.7} \\
\bottomrule
\end{NiceTabular*}
\end{table*}

\textbf{Robustness to Sparsity, Noise, and Frame Drop.}\label{sec,robustness} Section~\ref{sec,method} noted that NFM's local reference frame becomes ill-conditioned under sparse or degraded input. Table~\ref{tab,robustness} tests this directly by comparing MoSaiC with an NFM-ablated variant (CMSM + TDD + CTCP, \textit{i.e.}, the full model minus the NFM loss) under three perturbations applied consistently at pre-training, fine-tuning, and test time, reducing the number of points sampled per frame, adding Gaussian coordinate noise with $\sigma$ expressed in units of the unit-normalized scene extent, and randomly dropping frames. Applying the perturbation during pre-training is essential here, because the local PCA reference frame that defines the NFM target is computed only at pre-training time, a perturbation confined to the downstream stages could not test the conditioning of that frame at all. At the default operating point (2048 points, no noise, no dropped frames), NFM contributes $+1.9$ F1@10 and $+1.3$ mIoU on top of the ablated variant, matching the TDD-versus-TDD+NFM comparison in Table~\ref{tab,ablation_motion}. This contribution shrinks steadily as points are removed, down to $+0.2$ F1@10 at 256 points per frame, and similarly shrinks under frame dropping, down to $+0.6$ F1@10 at a 20\% drop rate, both perturbations directly degrade the point count and temporal adjacency that the local PCA reference frame depends on, exactly the failure mode anticipated in Section~\ref{sec,method}. In contrast, NFM's contribution is far less affected by Gaussian coordinate noise, retaining $+1.3$ to $+1.4$ F1@10 across $\sigma \in \{0.01, 0.02\}$ against the near-total loss observed under sparsity, since covariance-based PCA on a fixed, sufficiently large point set is inherently a smoothing operation and is far less sensitive to per-point jitter than to a reduction in the number of points available to estimate the covariance in the first place. Even where NFM's own contribution nearly vanishes, the full MoSaiC pipeline still degrades gracefully rather than catastrophically, because CMSM and CTCP do not share NFM's dependence on a well-conditioned local frame.

\begin{table*}[t]
\caption{Robustness of the NFM contribution under input degradation. w/o NFM denotes CMSM + TDD + CTCP without the NFM loss, Full denotes MoSaiC. $\Delta$ is the F1@10 gain attributable to adding NFM under each condition.}\label{tab,robustness}
\begin{tabularx}{\textwidth}{@{}>{\raggedright\arraybackslash}X ccc >{\columncolor{mosaicrow}}c >{\columncolor{mosaicrow}}c c@{}}
\toprule
\rowcolor{white}
Perturbation & Level & \makecell{w/o NFM\\F1@10} & \makecell{w/o NFM\\mIoU} & \cellcolor{white}\makecell{Full\\F1@10} & \cellcolor{white}\makecell{Full\\mIoU} & \makecell{$\Delta$\\F1@10} \\
\midrule
\multirow{4}{*}{Points per frame} & 2048 (default) & 84.3 & 42.4 & \textbf{86.2} & \textbf{43.7} & +1.9 \\
 & 1024 & 83.4 & 41.5 & \textbf{84.6} & \textbf{42.8} & +1.2 \\
 & 512 & 80.9 & 39.8 & \textbf{81.3} & \textbf{40.9} & +0.4 \\
 & 256 & 74.0 & 36.5 & \textbf{74.2} & \textbf{36.8} & +0.2 \\
\midrule
\multirow{2}{*}{Gaussian noise $\sigma$} & 0.01 & 83.8 & 41.9 & \textbf{85.1} & \textbf{43.0} & +1.3 \\
 & 0.02 & 81.0 & 40.0 & \textbf{82.4} & \textbf{41.2} & +1.4 \\
\midrule
\multirow{2}{*}{Frame drop rate} & 10\% & 83.0 & 41.2 & \textbf{84.3} & \textbf{42.6} & +1.3 \\
 & 20\% & 79.5 & 39.0 & \textbf{80.1} & \textbf{40.3} & +0.6 \\
\bottomrule
\end{tabularx}
\end{table*}

\textbf{Qualitative Results.} Figure~\ref{fig,qualitative} visualizes the complementary masking and reconstruction results of MoSaiC on representative action sequences, showing its ability to recover both the static structure and dynamic movements from the two masked views. \textbf{Feature Visualization.} To further assess the discriminativeness of the learned representations, Figure~\ref{fig,tsne} shows a t-SNE \citep{maaten2008tsne} projection of the encoder features on the action recognition test set. Compared with the from-scratch baseline, the features produced by MoSaiC form more compact and better-separated clusters per action category, qualitatively confirming that our pre-training learns more semantically structured representations.

\section{Discussion}\label{sec,discussion}

\begin{wrapfigure}{r}{0.48\linewidth}
  \centering
  \includegraphics[width=\linewidth]{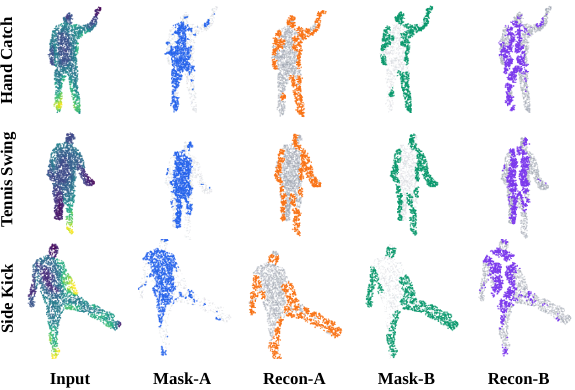}
  \caption{Qualitative visualization of complementary masking and reconstruction. For each sample we show the input, the two complementary masked views (Mask-A / Mask-B), and the corresponding reconstructions (Recon-A / Recon-B).}
  \label{fig,qualitative}
\end{wrapfigure}

\textbf{Robustness of the Motion Target.} NFM's supervision signal is derived from a local PCA reference frame, and Section~\ref{sec,method} anticipated that this frame becomes ill-conditioned under sparse or degraded input. Table~\ref{tab,robustness} confirms this is a real, quantifiable limitation rather than a theoretical concern, NFM's own contribution to F1@10 shrinks from $+1.9$ at the default 2048 points per frame to $+0.2$ at 256 points, and from $+1.9$ to $+0.6$ under a 20\% frame-drop rate, since both perturbations directly reduce the point count and temporal adjacency the local frame depends on. Encouragingly, the eigengap-based exclusion criterion we use to mask out unreliable tokens (Section~\ref{sec,method}) keeps this degradation graceful rather than catastrophic, because CMSM and CTCP do not share this dependency and continue to contribute under the same conditions. A confidence-weighted or fully learned formulation of the exclusion threshold, rather than our fixed one, is a natural next step for deployment on sensors with lower point density than the datasets studied here.

\textbf{Why the Margin on MSR-Action3D Is Smallest.} MoSaiC leads every baseline on all three action-recognition datasets, temporal action segmentation, semantic segmentation, label efficiency, linear probing, and cross-dataset transfer, but the margin it achieves on MSR-Action3D is the narrowest of the three action-recognition benchmarks, $+0.42$ over the strongest baseline, C2P (95.18 vs.\ 94.76, Table~\ref{tab,action_recognition}), compared with $+0.7$ on SHREC'17 and $+0.5$ on NvGesture. We attribute this to a mismatch between what NFM supervises and what the benchmark rewards, MSR-Action3D consists of 20 coarse, whole-body actions (\textit{e.g.}, \textit{high arm wave}, \textit{forward kick}) that are largely identifiable from posture in a handful of frames, so the fine-grained local rotation signal NFM provides has comparatively little additional discriminative value once appearance is well modeled. C2P's complete-to-partial distillation, in contrast, directly supervises whole-sequence completeness, which aligns more closely with a benchmark where the relevant motion is coarse and global rather than local and rotational. This is consistent with the pattern in Table~\ref{tab,ablation_components}, where NFM's increment is smaller on action recognition ($+0.8$) than on the two dense tasks ($+2.6$ F1@10, $+1.1$ mIoU), it is CTCP's instance-discriminative signal, not NFM's motion signal, that contributes most of MoSaiC's advantage on MSR-Action3D. We view this as an informative pattern that clarifies NFM's intended regime (local, dense, rotation-bearing motion) rather than as evidence against the general design.
\begin{figure}[htbp]
  \centering
  \includegraphics[width=\textwidth]{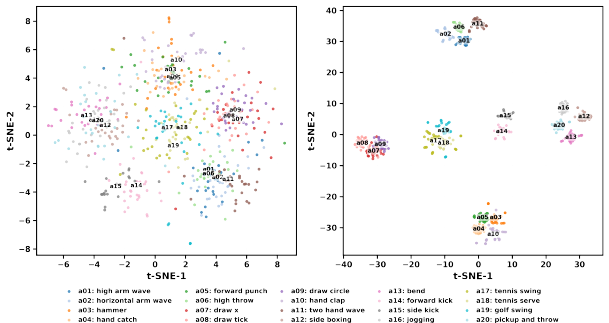}
  \caption{t-SNE visualization of the encoder features (each color denotes an action category). Left, training from scratch. Right, MoSaiC pre-training.}
  \label{fig,tsne}
\end{figure}
\section{Conclusion}\label{sec,conclusion}
In this paper, we presented MoSaiC, a self-supervised masked-modeling framework that learns transferable point cloud video representations by jointly exploiting motion-saliency-guided masking, explicit local geometric motion supervision, and complementary cross-view consistency. Experiments across four datasets and three downstream task families demonstrate its consistent effectiveness, particularly for dense spatio-temporal tasks and low-label settings, while highlighting robustness under severe input degradation as an important direction for future work.

\section*{CRediT authorship contribution statement}
\textbf{Wei Wang:} Conceptualization, Methodology, Writing -- original draft.
\textbf{Yiding Sun:} Methodology, Software, Validation, Writing -- original draft.
\textbf{Yuyan Wang:} Resources, Visualization.
\textbf{Zhuoyue Zhang:} Investigation.
\textbf{Zhengqiao Li:} Investigation.
\textbf{Dongfu Yin:} Writing -- review \& editing.
\textbf{Chen Li:} Writing -- review \& editing.

\section*{Declaration of competing interests}
The authors declare that they have no known competing financial interests or personal relationships that could have appeared to influence the work reported in this paper.

\section*{Acknowledgments}
This work was supported in part by the Technology Innovation Leading Program of Shaanxi (Program No. 2024QY-SZX-23), the Shenzhen Science and Technology Program under Grant KJZD20240903104400001, and the Research Task Assignment Project from Guangdong Laboratory of Artificial Intelligence and Digital Economy (SZ) under Grant No. GML-26420004, and by the High-performance Computing Platform of Peking University.

\bibliographystyle{elsarticle-num}
\bibliography{cas-refs}

\end{document}